# Sentiment Polarity Detection on Bengali Book Reviews Using Multinomial Naïve Bayes

Eftekhar Hossain[1], Omar Sharif[2] and Mohammed Moshiul Hoque[2°]

[1]Department of Electronics and Telecommunication Engineering
[2]Department of Computer Science and Engineering
Chittagong University of Engineering and Technology
Chittagong-4349, Bangladesh
{eftekhar.hossain, omar.sharif, moshiul_240°}@cuet.ac.bd

**Abstract** Recently, sentiment polarity detection has increased attention to NLP researchers due to the massive availability of customer's opinions or reviews in the online platform. Due to the continued expansion of e-commerce sites, the rate of purchase of various products, including books, are growing enormously among the people. Reader's opinions/reviews affect the buying decision of a customer in most cases. This work introduces a machine learning-based technique to determine sentiment polarities (either positive or negative category) from Bengali book reviews. To assess the effectiveness of the proposed technique, a corpus with 2000 reviews on Bengali books is developed. A comparative analysis with various approaches (such as logistic regression, naive Bayes, SVM, and SGD) also performed by taking into consideration of the unigram, bigram, and trigram features, respectively. Experimental result reveals that the multinomial Naive Bayes with unigram feature outperforms the other techniques with $84\%$ accuracy on the test set.

**Keywords:** Bangla language processing, Sentiment polarity detection, Feature extraction, Book reviews, Machine learning.

## 1 Introduction

Automatic detection of sentiment polarity is one of the notable research issues in opinion mining and natural language processing. The number of opinions or reviews on social media, blogs, and online platforms are growing enormously due to the substantial growth of internet uses and its uncomplicated access via e-devices. Online marketers facilitated their purchaser to convey feeling about the procured items or services to boost up purchaser contentment. The new purchasers prefer to read earlier posted product reviews, and they can make their decision to buy a particular item based on these reviews. Sentiment detection is a technique to estimate the point of view of a user on a specific matter. It categorizes the polarity of the text content (in the form of tweets, reviews, comments, posts, or bulletins), as the positive, neutral, or negative [12].

Recently, the purchasing habits of people are increased exponentially via online/e-commerce sites, and the book is one of the most selling products online. The amenities of the online book platform is that both the customers and authors can get a sight of how readers and reviewers responded to a particular book. Thus, book reviews may play an essential role in creating attraction or aversion on a particular book and help them in deciding to purchase this book. It may become a common scenario among the customers to read the reviews before buying a book. A book with an abundant amount of positive reviews can succeed in gaining the customer's attention and faith towards the publishers as well as authors. Nevertheless, it is a very complicated and time consuming task to scrutinize every review manually from a considerable amount of reviews. Thus this time-consuming task puts concealment for an automated method that preserve comprehend the contextual polarity of reviewer's annotations scripted at various social media groups as well as online book shops. An automatic sentiment polarity detection technique can use for better decision making, efficient manipulation, and understating customer feelings/opinion on a particular book. Sentiment analysis is associated with labeling a specific sentence with positive and negative annotation. The task of sentiment analysis is momentous to the business community, where user reviews need to take account to know the sus-tainability of their product and to identify the general feel of the consumers towards that product.

The sentiment polarity detection broadly classifies as sentiment knowledge-based and statistical-based [5]. In the sentiment knowledge-based approach, a sentiment dictionary is used to execute specific computation of words that bear sentiment in the content and figure out the exploration of sentiments from text [1]. On the other hand, the statistically-based approach picks a text analysis technique to extract appropriate linguistic features of the text and utilizes a machine learning technique to consider the sentiment analysis as a classification problem [20]. In our work, we use the statistical-based approach. As far as we concern, there is no study has been conducted to date on sentiment analysis from book reviews in the Bengali language. To address this issue, our contributions to this work can remark in the following:

– Develop a corpus consisting of 2000 Bengali book reviews, which are labeled as positive and negative sentiments.
– Develop a supervised machine learning model using Naive Bayes technique to categorize the sentiment from book reviews into positive or negative po-larities.
– Analyze the effect of several n-gram features on naive Bayes, logistic regres-sion, SVM and SGD algorithms by using develop dataset.
– Analyze the effectiveness of the proposed technique on different distributions of the developed dataset.
– Assess the outcome of the proposed technique by comparing with the other machine learning techniques on the developed dataset.

The remaining of the paper is arranged following: Section 2 represents related work. A brief description of the developed corpus is outlined in Section 3. Section

4 provides the suggested framework with a detailed explanation of its constituent parts. Section 5 states the experimentation with the analysis of findings. Finally, the paper is concluded with a summary in Section 6.

## 2 Related Work

Sentiment analysis or detection from text is a well-studied research issue for highly resourced languages like English, Arabic, and other European languages [10]. Srujan et al. [15] presented a random forest technique to analyze sentiment of amazon book reviews. They used TF-IDF to extracts n-gram features and obtained the accuracy of about 90.15%. Akshay et al. [9] explored machine learning classifiers for analyzing sentiment of the restaurant reviews. They have obtained the highest accuracy of 94.5% for their dataset. A positive and negative sentiments detection model is developed on cell phone reviews using SVM, which achieved an accuracy of 81.77% [14]. Fang et al. [4] presented a model for analyzing sentiment on online product reviews where a sentiment score formula along with three classifiers was used for the categorization of text polarity. Chinese text-based sentiment analysis is developed using Naive Bayes, where a sentiment dictionary is used for classification [18]. A classifier model is proposed in the Arabic language for grouping reviews of social networks [7]. A deep learning-based emotion detection technique is developed by Xu et al. [19] on the medical/health dataset, which achieved 85% accuracy on emotional fatigue.

Since Bengali is an under-resourced language, the amount of e-text contents or reviews in the Bengali language (primarily related to books) is quite limited. In addition to that, no benchmark dataset is available on sentiment classification of book reviews in the Bengali language. Due to these barriers, very few research activities have been conducted in this area of Bangla language processing (BLP), which are mainly related to sentiment detection from news, restaurant reviews, product reviews, social media reviews, micro-blogging comments, and so on. Rumman et al. [2] proposed a sentiment classifier for movie reviews written in Bengali text. They have explored different machine learning techniques over a small review dataset. Their dataset shows excellent performance for SVM and LSTM model with an accuracy of 88.90% and 82.42%, respectively. An SVM based sentiment analysis on the Banglaseh Cricket dataset is developed, which achieved 64.59% accuracy [11]. Sarkar et al. [13] presented a sentiment classification system on the Bengali tweet dataset, where SVM and multinomial naive Bayes classifiers used for the classification task. Their system achieved the highest 45% accuracy for SVM classifier over n-gram and SentiWordNet features. A recent technique to analyze sentiment from the Bengali text was proposed by Taher et al. [16], where various n-gram features incorporated with the SVM classifier. They have obtained maximum accuracy of 91.68% for linear SVM. A supervised model proposed to determine the positive and negative sentiments from Facebook status written in Bengali [8]. This system achieved 0.72 f-score using a multinomial Naive Bayes classifier with bigram features. Another model of sentiment analysis on the Bengali horoscope corpus is proposed based on SVM, which achieved 98.7% accuracy using unigram features [6]. It observed that most of the research activities have been conducted so far, considered a small dataset of Bengali sentiment analysis. Furthermore, these considered movie reviews and social blog comments for sentiment analysis. As far as we are aware that none of the work has done yet on sentiment analysis on Bengali book reviews. This work proposes a technique for detecting positive and negative sentiments from online book reviews written in Bengali.

## 3   Dataset Preparation

Bengali is one of the resource-poor languages due to its insufficient e-contents and unavailability of the standard dataset. To serve our purpose, we developed our dataset on sentiment polarity by collecting data from the available web resources such as blogs, Facebook, and e-commerce sites. We endorsed a technique of developing a dataset, as explained by Das et al. [3].

Table 1. Sample dataset.

| Sample Reviews | Sentiment |
|---|---|
| অতি অসাধারণ একটা ডার্ক হরর। এক নিঃশ্বাসে পড়ে শেষ করার মতো বই। খুবিই ভালো। (An extraordinary dark horror. This is a breath taking book. That's good) | Positive |
| এটা আসলে বই ছিল না এক ধরনের বাজে রসিকতা তা বোধহয় সমরেশ মজুমদার বলতে পারবেন। অখাদ্য। পুরোই মেজাজটাই খারাপ হয়ে গেল। (Only shomresh majumder can tell, was it really a book or a kind of crap joke. Disgusting. The whole mood went bad..) | Negative |

Table 1 shows a negative and positive review as examples. The developed dataset contained 2000 textbook reviews, and these reviews randomly divided in 3 distributions: training set ($T_R$), validation set ($V_D$), and test set ($T_S$), respectively. Table 2 exhibits the summary of the developed data set.

**Table 2.** Summary of the dataset.

| Dataset attributes | $T_R$ | $V_D$ | $T_S$ |
|---|---|---|---|
| Number of documents | 1600 | 200 | 200 |
| Number of words | 29079 | 4458 | 5234 |
| Total unique words | 8336 | 1193 | 1518 |
| Size (in bytes) | 974848 | 79872 | 70656 |
| Number of sentences | 6728 | 1137 | 876 |

## 4 Sentiment Polarity Detection Framework

Fig. 1 depicts the suggested technique for sentiment polarity detection/classification. This framework composes of three major phases: preprocessing, feature extraction, and classification, respectively.

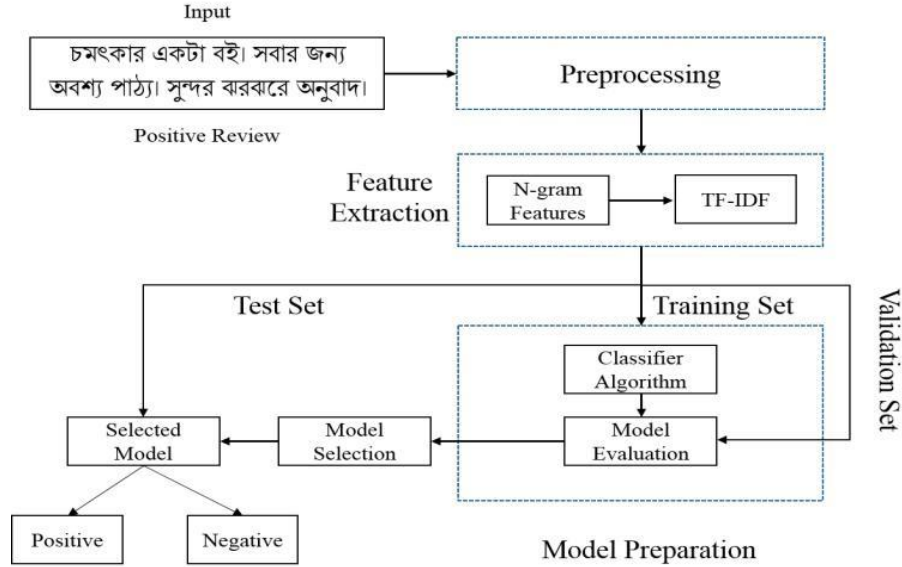

**Fig. 1.** Proposed framework for sentiment polarity detection.

### 4.1 Preprocessing

Text reviews $r^{(i)}$ of the corpus $\mathcal{R}[]$ can processes in several steps. To illustrate the pre-processing steps a sample review **r** = " ইহা এক অসাধারণ বই ।। ...!! " have selected from the corpus.

1. **Removal of redundant characters:** punctuation symbols, special characters and numbers are removed from the collected reviews. These are con-sidered redundant as they do not bear any sentiment information. After removing the redundant characters the sample review becomes **r**=[ইহা এক অসাধারণ বই ]
2. **Tokenization:** is the process of partitioning a review into it's constituent words. We get a vector of words $r = \{w^{<1>}, w^{<2>}, ..., w^{<l>}\}$ by tokenizing a review $r$ of $l$ words. **r** =['ইহা', 'এক ',' অসাধারণ' ,'বই' ].
3. **Stop words removal:** A word $w^{<i>}$ has been removed from a review $r$ by matching it with a stop word list $S[] = \{s_1, s_2, ..., s_t\}$ having $t$ stop words. A word $w^{<i>}$ which did not contribute to fix whether a review is positive or negative has been considered as a stop word. Such as conjunction, pronoun, preposition etc. At this step the sample review becomes **r** =[অসাধারণ বই ].

A clean corpus has been developed by performing all prepossessing steps on reviews.

### 4.2 Feature Extraction Techniques

The feature extraction process determines the degree of success of any machine learning-based system. N-gram features of the text have experimented with the proposed approach. Table 3 shows unigram, bigram, and trigram features for a sample review (লেখকের উপস্থাপনা বেশ চমৎকার) with total system features.

Table 3. Illustration of various N-grams features.

| Sample Text | লেখকের উপস্থাপনা বেশ চমৎকার | Total Features |
|---|---|---|
| Uni-gram feature | 'লেখকের', 'উপস্থাপনা', 'বেশ', 'চমৎকার' | 2364 |
| Bi-gram feature | 'লেখকের উপস্থাপনা', 'উপস্থাপনা বেশ', 'বেশ চমৎকার' | 14612 |
| Tri-gram feature | 'লেখকের উপস্থাপনা বেশ', 'উপস্থাপনা বেশ চমৎকার' | 19588 |

A common way of extracting "*tf- idf*" feature-value from a text is computed by the equation 1 [17].

$$tfidf(w^{<i>}, r^{(i)}) = tf(w^{<i>}, r^{(i)}) \log \frac{N}{|\{r \epsilon \Re : w \epsilon r\}|} \quad (1)$$

where, *tfidf(w, r)* represents the value of word $w^{<i>}$ in review $r^{(i)}$, *tf(w, r)* means the frequency of $w^{<i>}$ in review $r^{(i)}$, N represents the total number of reviews, |{r∈ℜ : w∈r}| indicates the number of reviews containing $w^{<i>}$.

The combination of n-gram features and *tf idf* values were used as the features of the proposed system. Table 4 represents a small fragment of *tfidf* values for n-gram features for some arbitrary reviews.

Table 4. A small fragment of feature space.

| ℜ, C | Uni-gram | | | Bi-gram | | | Tri-gram | | |
|---|---|---|---|---|---|---|---|---|---|
| | $w_1$ | .. | $w_u$ | $w_1$ | .. | $w_b$ | $w_1$ | .. | $w_t$ |
| $r^{(1)}$ | 0.45 | .. | 0.231 | 0.31 | .. | 0.431 | 0.57 | .. | 0.35 |
| $r^{(2)}$ | 0.25 | .. | 0.51 | 0.24 | .. | 0.253 | 0.53 | .. | 0.57 |
| $r^{(3)}$ | 0.63 | .. | 0.09 | 0.13 | .. | 0.356 | 0.20 | .. | 0.46 |
| $r^{(4)}$ | 0.17 | .. | 0.12 | 0.65 | .. | 0.134 | 0.036 | .. | 0.35 |
| $r^{(5)}$ | 0.29 | .. | 0.32 | 0.37 | .. | 0.541 | 0.076 | .. | 0.35 |

Features are represented in a two dimensional space where reviews $\mathcal{R} = r^{(1)}, r^{(2)}, ..., r^{(n)}$ in rows and n-gram features are represented in columns. Columns are divided into 3 parts and these are (i) Uni-gram feature set $U = \{w_1, w_2, ..., w_u\}$ (ii) Bi-gram feature set $B = \{w_1, w_2, ..., w_b\}$ (iii) Tri-gram feature set $T = \{w_1, w_2, ..., w_t\}$. The number in each cell holds the *tfidf* value of word $w_i$ in the feature set $\{U, B, T\}$ for a review $r_i$. Each row correspond to the feature vector $F = F_V[1], F_V[2], F_V[3], ..., F_V[n]$ for reviews $\mathcal{R} = r^{(1)}, r^{(2)}, r^{(3)}, ..., r^{(n)}$ of the corpus.

### 4.3 Classifier Model Preparation

The proposed work aims to develop a sentiment classifier to categorize a book review either in positive or negative category. N-gram features with *tf idf* (de-scribed in Subsection 4.2) mainly used for the model preparation. The extracted features segmented into three random distributions, and each distribution applied to a particular stage of the model preparation, such as classification and model evaluation.

**Classification** In this stage a set of classifier models $M[] = \{m_1, m_2, ..., m_7\}$ has been developed by applying different learning algorithms on the training set $T_R = \{r^{(1)}, r^{(2)}, ..., r^{(x)}\}$. These algorithms are logistic regression (LR), decision tree (DT), random forest (RF), multinomial naive Bayes (MNB), KNN, SVM and stochastic gradient descent (SGD) respectively.

**Model evaluation** For the purpose of evaluation 10-fold cross-validation have been performed on each of the model $M[i]$. Among these models, the best four (LR, SVM, MNB, and SGD) will be chosen based on their cross-validation accuracy, which is done by algorithm 1. These models will be evaluated on

---

**Algorithm 1** Finding best 4 models based on cross validation scores.

1: Initialize $CV_m = [], i = 0$;   "$CV_m$ = Set of cross validates model"
2: Initialize $CV_s = [s_1, s_2, ..., s_7]$;   "$CV_s$ = List of cross validation score"
3: Sort($CV_s$, $CV_s + 7$)
4: $CV_m.append(CV_s[i])$
5: **if** $i == 4$ **then**
6:    *exit*
7: **else**
8:    $i = i + 1$
9:    *go to step* 4
10: **end if**

---

different parameters by using the validation set $V_D = \{r^{(1)}, r^{(2)}, ..., r^{(y)}\}$ to find out the desired one. Finally test set $T_S = \{r^{(1)}, r^{(2)}, ..., r^{(z)}\}$ will be used to assess the selected model.

## 5  Results Analysis

The performance assessment of the proposed technique is performed by using several graphical and statistical measures such as confusion matrix, $f_1$ score, recall, precision, ROC, and precision versus recall curve. For the development of the sentiment classification model initially, seven classifiers have been selected for the training. 10- fold cross-validation has been done over all the classifiers using n-gram features. The trained classifiers are LR, MNB, RF, DT, KNN, SVM and SGD. Among these seven trained classifier models, only four classifiers provide acceptable cross-validation accuracy and thus selected them for further evaluation over validation data. Table 5 shows the 10-fold cross-validation ac-curacy over the n-gram features for all the classifiers. The result indicates that that KNN, DT, and RF algorithms achieved the lower accuracy for all cases of n-gram features.

**Table 5.** 10-Fold cross validation results.

|  | Classifier | Uni-gram | Bi-gram | Tri-gram |
|---|---|---|---|---|
|  | LR | **83** | 77 | 66 |
|  | KNN | 58 | 54 | 60 |
|  | DT | 69 | 77 | 59 |
| Accuracy(%) | RF | 73 | 77 | 68 |
|  | MNB | **88** | 78 | 69 |
|  | SVM | **81** | 77 | 63 |
|  | SGD | 77 | 72 | 68 |

Selected four classifier models assessed by using the validation dataset. Table 6 illustrates the performance of four classifiers in terms of accuracy, precision, re-call, and $f_1$ measures. It observed that in the case of the uni-gram feature, MNB achieved the highest accuracy of about 87%. For bi-gram features, both LR and SGD achieved the highest accuracy of about 80%, whereas MNB provided the lowest accuracy (74%). On the contrary, for tri-gram features, all four classi-fiers provided the lowest accuracy (only 70%) than the bi-gram and tri-gram features. Although the number of features increased in tri-gram, the limited number of reviews in each class suppressed the chances of occurring features in several reviews. As a result, the overall accuracy decreased. The result of the analysis shows that multinomial naive Bayes for unigram features outperformed the other classifiers and features extraction methods.

For all classifiers, the experiment performed again for graphical analysis. Fig. 2 to Fig. 4 shows ROC and PR curve for the selected four classifier models with unigram, bigram, and trigram features, respectively. In the case of unigram features, MNB provides the highest AUC at 89.6% and average precision at

**Table 6.** Performance measures of classifiers.

| Features | Classifier | Accuracy(%) | Precision(%) | Recall(%) | $f_1$ score(%) |
|---|---|---|---|---|---|
| Uni-gram | LR | 81 | 81 | 81 | 81 |
| | MNB | **87** | **89** | 86 | 86 |
| | SVM | 81 | 81 | 81 | 81 |
| | SGD | 78 | 82 | 76 | 76 |
| Bi-gram | LR | 80 | 81 | 81 | 80 |
| | MNB | 74 | 84 | 71 | 70 |
| | SVM | 76 | 81 | 78 | 76 |
| | SGD | 80 | 84 | 82 | 80 |
| Tri-gram | LR | 61 | 60 | 95 | 73 |
| | MNB | 64 | 61 | 95 | 75 |
| | SVM | 68 | 89 | 49 | 63 |
| | SGD | 60 | 58 | 95 | 73 |

91.2%. On the other hand, for bigram features, all the classifiers jointly give good AUC and AP score for ROC and PR curves, respectively.

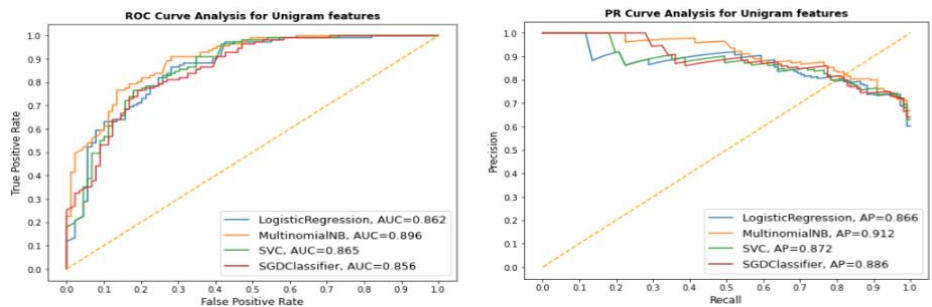

**Fig. 2.** Classifiers performance on Uni-gram features.

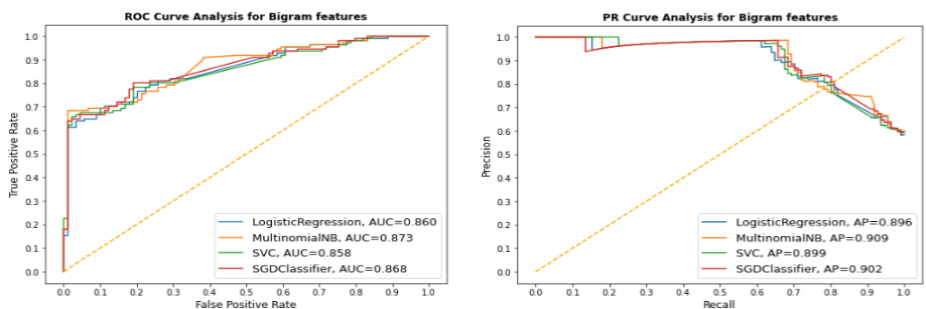

**Fig. 3.** Classifiers performance on Bi-gram features.

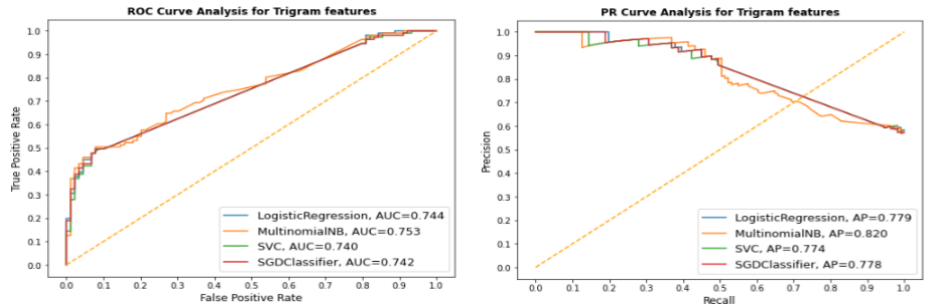

**Fig. 4.** Classifiers performance on Tri-gram features.

In the case of trigram features, the AUC and AP scores show the lowest score for every classifier model. The low AUC and AP scores reported by the classifiers imposed the fact that in trigram features. Many of the strong sentiment carrying words and phrases misrepeated in most of the reviews.

Detailed analysis of the results brings the notation that the MNB with uni-grams features gives the highest accuracy, AUC, and AP scores in comparison with all the other classifier models and gram features. Thus, a multinomial naive Bayes classifier selected as the final model. Table 7 depicts the detailed classification reports of the MNB on the test set.

**Table 7.** Summary of evaluation of the proposed technique.

| Sentiment polarity ($C$) | Precision | Recall | $f_1$-score | Support |
|---|---|---|---|---|
| Negative ($C_n$) | 0.91 | 0.72 | 0.81 | 89 |
| Positive ($C_p$) | 0.81 | 0.95 | 0.87 | 111 |
| avg./total | 0.86 | 0.83 | 0.84 | 200 |

From the classification report, it observed that the value of $f_1$-score is $0.81$ and $0.91$ respectively for positive and negative sentiment class. From the analysis of the results, it concluded that the proposed model successfully classifies $91\%$ of negative, and $81\%$ of positive book reviews and achieved 84% overall accuracy on the test set.

The overall classification performance of the proposed system is oscillating between 60% to 85%. It is due to the lack amount of training samples, and it can be improved further by including more data on the training phase. Fig. 5 shows the impact of the number of training samples on the model performance. This figure reveals that the accuracy increases with the increase in training dataset for unigram features.

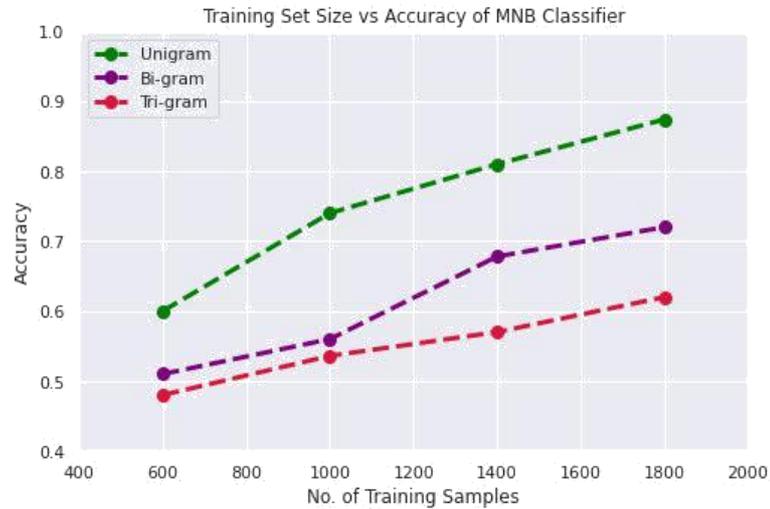

**Fig. 5.** Effect of classifier performance on training sets.

## 6 Conclusion

This paper presents a machine learning-based sentiment classification framework that can determine the sentiment into positive and negative categories from the Bengali book reviews by exploiting the various feature extraction techniques. The combination of $tf \cdot idf$ values with n-gram features considered as the best feature of the proposed framework. These extracted features are applied to clas-sify the inherent sentiment of the Bangla text reviews using various ML tech-niques. Among these classifiers, multinomial naive Bayes with unigram features provided the highest accuracy of 87% and 84% for validation and test datasets, respectively. Analyzing the sentiment of a book is very helpful for the authors and publishers as it can aid them by providing an abstract that what readers think about a book. In the future, deep learning techniques and sophisticated feature extraction methods such as word2Vec or Glove may apply for improved accuracy.